\documentclass[fleqn] {article} % For LaTeX2e
\usepackage{nips11submit_e,times}
\usepackage{graphicx}
\usepackage{epsfig}
\usepackage{amsmath}
\usepackage{amssymb}
\usepackage{mathrsfs}
\usepackage{algorithmic}
\usepackage{algorithm}
\usepackage{multirow}
\usepackage{booktabs}
\usepackage{caption}
\usepackage{subcaption}
\usepackage[pagebackref=true,breaklinks=true,letterpaper=true,colorlinks,bookmarks=false]{hyperref}

\title{Dynamical And-Or Graph Learning for Object Shape Modeling and Detection}

\author{
Xiaolong Wang\\
Sun Yat-Sen University\\
Guangzhou, P.R. China 510006 \\
\texttt{dragonwxl123@gmail.com} \\
\And
Liang Lin\thanks{Corresponding author is Liang Lin. This work was supported by National Natural Science Foundation of China (no. 61173082), Fundamental Research Funds for the Central Universities (no. 2010620003162041), and the Guangdong Natural Science Foundation (no.S2011010001378).This work was also partially funded by SYSU-Sugon high performance computing typical application project.} \\
Sun Yat-Sen University \\
Guangzhou, P.R. China 510006\\
\texttt{linliang@ieee.org} \\
}

\nipsfinalcopy % Uncomment for camera-ready version

\begin{document}

\maketitle

\begin{abstract}

This paper studies a novel discriminative part-based model to represent and recognize object shapes with an ``And-Or graph''. We define this model consisting of three layers: the leaf-nodes with collaborative edges for localizing local parts, the or-nodes specifying the switch of leaf-nodes, and the root-node encoding the global verification. A discriminative learning algorithm, extended from the CCCP~\cite{CCCP}, is proposed to train the model in a dynamical manner: the model structure (e.g., the configuration of the leaf-nodes associated with the or-nodes) is automatically determined with optimizing the multi-layer parameters during the iteration. The advantages of our method are two-fold. (i) The And-Or graph model enables us to handle well large intra-class variance and background clutters for object shape detection from images. (ii) The proposed learning algorithm is able to obtain the And-Or graph representation without requiring elaborate supervision and initialization.  We validate the proposed method on several challenging databases (e.g., INRIA-Horse, ETHZ-Shape, and UIUC-People), and it outperforms the state-of-the-arts approaches.
\end{abstract}
\vspace{-2ex}

\section{Introduction}
\vspace{-2ex}
Part-based and hierarchical representations have been widely studied in computer vision, and lead to some elegant frameworks for complex object detection and recognition. However, most of the methods address only the hierarchical decomposition by tree-structure models~\cite{LatentSVM,LeoCCCP}, and oversimplify the reconfigurability (i.e. structural switch) in hierarchy, which is the key to handle the large intra-class variance in object detection. In addition, the interactions of parts are often omitted in learning and detection. And-Or graph models are recently explored in \cite{LeoAOG,AOGgrammar} to hierarchically model object categories via ``and-nodes'' and ``or-nodes'' that represent, respectively, compositions of parts and structural variation of parts. Their main limitation is that the learning process is strongly supervised and the model structure needs to be manually annotated.

The key contribution of this work is a novel And-Or graph model, whose parameters and structure can be jointly learned in a weakly supervised manner. We achieve the superior performance on the task of detecting and localizing shapes from cluttered backgrounds, compared to the state-of-the-art approaches. As Fig.~\ref{fig:exp1}(a) illustrates, the proposed And-Or graph model consists of three layers described as follows.

The \textbf{leaf-nodes} in the bottom layer represent a batch of local classifiers of contour fragments. We provide a partial matching scheme that can recognize the accurate part of the contour, to deal with the problem that the true contours of objects are often connected to background clutters due to unreliable edge extraction.

The \textbf{or-nodes} in the middle layer are ``switch'' variables specifying the activation of their children leaf-nodes. We utilize the or-nodes accounting for alternate ways of composition, rather than just defining multi-layer compositional detectors, which is shown to better handle the intra-class variance and inconsistency caused by unreliable edge detection. Each or-node is used to select one contour from the candidates detected via the associated leaf-nodes in the bottom layer. Moreover, during detection, location displacement is allowed for each or-node to tackle the part deformation.

The \textbf{root-node} (i.e. the and-node) in the top layer is a global classifier capturing the holistic deformation of the object. The contours selected via the or-nodes  are further verified as a whole, in order to make the detection robust against the background clutters.

The \textbf{collaborative edges} between leaf-nodes are defined by the probabilistic co-occurrence of local classifiers, which relax the conditional independence assumption commonly used in previous tree structure models. Concretely, our model allows nearby contours to interact with each other.

The key problem of training our And-Or graph model is automatic structure determination. We propose a novel learning algorithm, namely dynamic CCCP , extended from the concave-convex procedure (CCCP)~\cite{CCCP,SVMICML2009} by embedding the structural reconfiguration. It iterates to dynamically determine the production of leaf-nodes associated with the or-nodes, which is often simplified by manually fixing in previous methods~\cite{LeoCCCP,ShiShapeCVPR2010}. The other structure attributes (e.g., the layout of or-nodes and the activation of leaf-nodes) are implicitly inferred with the latent variables.

%The other structure attributes (e.g., the layout of or-nodes and the activation of leaf-nodes) are implicitly decided by model parameters, which can be learned by maximizing the discrimination between object shape and background clutters.

%, which can be learned by maximizing the discrimination between object shape and background clutters.

%The inference of shape detection is consistent with the model training, including two steps: i) a bottom-up stage to make proposals via the leaf-nodes (i.e., detecting contours) and the or-nodes(i.e., selecting contours); ii) a top-down stage to validate the proposals via the root-node and the collaborative edges. Specifically, the selected contours are combined together for global verification where their contextual interactions are encoded.

\vspace{-1mm}
\section{Related Work}
\vspace{-2ex}
%old version

%The learning of hierarchical model for object detection has been an active area of research, such as generative learning of active basis~\cite{ActiveBasis}, deep learning~\cite{DBN}, random forest~\cite{HoughForest,Kinect} and bag of words~\cite{BagOfWords}. Recently, the discriminative part-base models learned via latent-SVM ~\cite{LeoCCCP,ShiShapeCVPR2010,LatentSVM,ContextSVMCVPR2010} have provided significant improvements on object detection. These methods studied on the hierarchical tree structure model encoding how sub-parts are composed into a larger structure. However, they assume that the local parts on the same layer are independent of each other, which is somewhat restrictive. With the introduction of the hidden conditional random field(HCRF) ~\cite{HCRF}, this assumption is relaxed. Inspired by this work, ~\cite{MMHCRF} proposes a discriminative part-base model with HCRF, allowing nearby parts to interact with each other. And ~\cite{HierachicalCVPR2009} also presents a hierarchical multi-feature representation of each object as a CRF with a pairwise graph structure. To improve the graph model that not only encoding the interplay of different parts, but also capturing the large intra-class variance, the And-Or graph model is proposed in~\cite{AOGgrammar}. Despite its manually defined structure and supervised-learning, the And-Or graph model has been applied widely in object detection and human parsing~\cite{LeoAOG,HierarchicalPoslets}, event understanding ~\cite{EventGrammar} and scene parsing~\cite{SceneParsing}.
%

Remarkable progress has been made in shape-based object detection ~\cite{FerrariIJCV09,LateckiCVPR2011,ShapeGroup,MalikCVPR2009,LateckiECCV2010}. By employing some shape descriptors and matching schemes, many works represent and recognize object shapes as a loose collection of local contours. For example, Ferrari et al.~\cite{FerrariIJCV09} used a codebook of PAS (pairwise adjacent segments) to localize object of interest; Maji et al.~\cite{MalikCVPR2009} proposed a maximum margin hough voting for hypothesis regions combining with intersection kernel SVM(IKSVM) for verification; Yang and Latecki~\cite{LateckiECCV2010} constructed shape models in a fully connected graph form with partially-supervised learning, and detected objects via a Particle Filters (PF) framework.

Recently, the tree structure latent models~\cite{LeoCCCP,LatentSVM} have provided significant improvements on object detection. Based on these methods, Srinivasan et al.~\cite{ShiShapeCVPR2010} trained the descriptive contour-based detector by using the latent-SVM learning; Song et al.~\cite{ContextSVMCVPR2010} integrated the context information with the learning, namely Context-SVM. Schnitzspan et al.~\cite{HierachicalCVPR2009} further combined the latent discriminative learning with conditional random fields using multiple features.

Knowledge representation with And-Or graph was first introduced for modeling visual patterns by Zhu and Mumford~\cite{AOGgrammar}. Its general idea, i.e. using configurable graph structures with And, Or nodes, has been applied in object and scene parsing~\cite{LeoAOG,HierarchicalPoslets,SceneParsing} and action classification ~\cite{FeifeiICML}.

%\section{And-Or Graph Representation}
%
%1. Leaf-nodes: local classifier
%2. Horizontal edges among leaf-nodes.
%3. Or-nodes: switch the compositions.
%4. The and-node: top-down verification.

\vspace{-1mm}
\section{And-Or Graph Representation for Object Shape}
\vspace{-2ex}
The And-Or Graph model is defined as $\mathcal{G}= (\mathcal{V},\mathcal{E})$, where $\mathcal{V}$ represents three types of nodes and $\mathcal{E}$ the graph edges. As Fig.~\ref{fig:exp1}(a) illustrates, the square on the top is the root-node representing the complete object instances. The dashed circles derived from the root are $z$ or-nodes arranged in a layout of $b_1 \times b_2$ blocks, representing the object parts. Each or-node comprises an unfixed number of leaf-nodes (denoted by the solid circles on the bottom); the leaf-nodes are allowed to be dynamically created and removed during the learning. For simplicity, we set the maximum number $m$ of leaf-nodes affiliated to one or-node, and the parameters of non-existing leaf-nodes to zero. Then the maximum number of all nodes in the model is $ 1 + n = 1 + z + z \times m$. We use $i = 0$ indexing the root node, $i=1,...,z$ the or-nodes and $j = z+1,...,n$ the leaf-nodes. We also define that $j \in ch(i)$ indexes the child nodes of node $i$. The horizontal graph edges (i.e., collaborative edges) are defined between the leaf-nodes that are associated with different or-nodes, in order to encode the compatibility of object parts. The definitions of $\mathcal{G}$ are presented as follows.

\textbf{Leaf-node:} Each leaf-node $L_j,j=z + 1,...,n$ is a local classifier of contours, whose placement is decided by its parent or-node (the localized block). Suppose a contour fragment $c$ on the edge map $X$ is captured by the block located at $p_i = (p_i^x,p_i^y)$, as the input of classifier. We denote $\phi^{l}(p_i,c)$ as the feature vector using the Shape Context descriptor~\cite{ShapeContext}. For any classifier, only the part of $c$ fallen into the block will be taken into account, and we set $\phi^{l}(p_i,c) = 0$ if $c$ is entirely out.
%Note that only the part of $c$ inside the block is under consideration, thus we can distinguish the true object contour from the background clutters linked to it. To be more specific, if the contour $c$ is entirely out of the block, $\phi^{l}(p_i,c) = 0$.
The response of classifier $L_j$ at location $p_i$ of the edge map $X$ is defined as:
\vspace{-1ex}
\begin{eqnarray}
&& \mathcal{R}_{L_j}(X,p_i) =\max_{c \in X} \omega_{j}^{l} \cdot \phi^{l}(p_i,c),
\end{eqnarray}
\vspace{-4.5ex}

where $\omega_{j}^{l}$ is a parameter vector, which is set to zero if the corresponding leaf-node $L_j$ is nonexistent. Then we can detect the contour from edge map $X$ via the classifier, $c_j = argmax_{c \in X} \omega_{j}^{l} \cdot \phi^{l}(p_i,c)$.

\textbf{Or-node:} Each or-node $U_i,i=1,...,z$ is proposed to specify a proper contour from a set of candidates detected via its children leaf-nodes. Note that we can also consider the or-node activating one leaf-node. The or-nodes are allowed to perturb slightly with respect to the root. For each or-node $U_i$, we define the deformation feature as $\phi^{s}(p_0,p_i)=(dx,dy,dx^2,dy^2)$, where $(dx,dy)$ is the displacement of the or-node position $p_i$ to the expected position $p_0$ determined by the root-node. Then the cost of locating $U_i$ at $p_i$ is:

\vspace{-4.5ex}
\begin{eqnarray}
&& Cost_{i}(p_0,p_i) = -\omega_{i}^{s} \cdot \phi^{s}(p_0,p_i),
\end{eqnarray}
\vspace{-4.5ex}

where $\omega_{i}^{s}$ is a 4-dimensional parameter vector corresponding to $\phi^{s}(p_0,p_i)$. In our method, each or-node contains at most $m$ leaf-nodes, among which one is to be activated during inference. For each leaf-node $L_j$ associated with $U_i$, we introduce an indicator variable $v_{j} \in \{0,1\} $ representing whether it is activated or not. Then we derive the auxiliary ``switch'' vector for $U_i$, $\textbf{v}_{i} = (v_{j_1},v_{j_2},...,v_{j_m} )$, where $ ||\textbf{v}_{i}|| = 1$. Thus, the response of the or-node $U_i$ is defined as,

\vspace{-4.5ex}
\begin{eqnarray} \label{eq:or-node_score}
&& \mathcal{R}_{U_i}(X,p_0,p_i,\textbf{v}_{i}) = \sum_{j \in ch(i)} \mathcal{R}_{L_j}(X,p_i) \cdot v_j + Cost_{i}(p_0,p_i).
\end{eqnarray}
\vspace{-4ex}

\textbf{Collaborative Edge:} For any pair of leaf-nodes $(L_j, L_{j^{\prime}})$ respectively associated with two different or-nodes, we define the collaborative edge between them according to their contextual co-occurrence. That is, how likely it is that the object contains contours detected via the two leaf-nodes. The response of the pairwise potentials is parameterized as,

\vspace{-4.5ex}
\begin{eqnarray}\label{eq:CRF score}
&& \mathcal{R}_{E}(V) = \sum_{j=z+1}^{n} \sum_{j^{\prime} \in neigh(j)} \omega_{(j,j^{\prime})}^{e} \cdot v_j \cdot v_{{j}^{\prime}},
\end{eqnarray}
\vspace{-4ex}

where $neigh(j)$ is defined as the neighbor leaf-nodes from the other or-node adjacent (in spatial direction) to $L_j$, and $V$ is a joint vector for each $\textbf{v}_i$: $V =(\textbf{v}_1,...,\textbf{v}_{z}) = (v_{z+1},...,v_n)$. $\omega_{(j,j^{\prime})}^{e}$ indicates the compatibility between $L_j$ and $L_{j^{\prime}}$.

\textbf{Root-node:} The root-node represents a global classifier to verify the ensemble of contour fragments $C^r=\{c_1,...,c_z\}$ proposed by the or-nodes. The response of the root-node is parameterized as,

\vspace{-4.5ex}
\begin{eqnarray} \label{eq:root_score}
&& \mathcal{R}_{T}(C^r) = \omega^{r} \cdot \phi^{r}(C^r),
\end{eqnarray}
\vspace{-4.5ex}

where $\phi^{r}(C^r)$ is the feature vector of $C^r$ and $\omega^{r}$ the corresponding parameter vector.

Therefore, the overall response of the And-Or graph is:

\vspace{-4.5ex}
\begin{small}
\mathindent=0pt
\begin{align} \label{eq:AOG_score}
& \qquad\qquad\quad\quad \mathcal{R}_{G}(X,P,V) =  \sum_{i=1}^{a} \mathcal{R}_{U_i}(X,p_0,p_i,\textbf{v}_{i}) + \mathcal{R}_{E}(V) + \mathcal{R}_{T}(C^r)\nonumber\\
& = \sum_{i=1}^{z}[\sum_{j \in ch(i)} \omega_{j}^{l} \cdot \phi^{l}(p_i,c_j) \cdot v_j - \omega_{i}^{s} \cdot \phi^{s}(p_0,p_i)] +
 \sum_{j=z+1}^{n} \sum_{j^{\prime} \in neigh(j)} \omega_{(j,j^{\prime})}^{e} \cdot v_j \cdot v_{{j}^{\prime}} + \omega^{r} \cdot \phi^{r}(C^r),
\end{align}
\mathindent=\leftmargini
\end{small}
\vspace{-3ex}

where $P=(p_0,p_1,...,p_z)$ is a vector of the positions of or-nodes. For better understanding, we refer $H = (P,V)$ as the latent variables during inference, where $P$ implies the deformation of parts represented by the or-nodes and $V$ implies the discrete distribution of leaf-nodes (i.e., which leaf-nodes are activated for detection). The Eq.(\ref{eq:AOG_score}) can be further simplified as :

\vspace{-4.5ex}
\begin{eqnarray}\label{eq:sim_score_infer}
&& \mathcal{R}_{G}(X, H) = \omega \cdot \phi(X,H),
\end{eqnarray}
\vspace{-4.5ex}

where $\omega$ includes the complete parameters of And-Or graph, and $\phi(X,H)$ is the feature vector,

\vspace{-4ex}
\begin{small}
\begin{align} \label{abteq65}
& \omega = (\omega_{z+1}^{l},...,\omega_{n}^{l},-\omega_{1}^{s},...,-\omega_{z}^{s}, \omega_{(z+1,{z+1+m})}^{e},..., \omega_{(n-m,n)}^{e} , \omega^{r}). \\
& \phi(X,H) = (\phi^{l}(p_1,c_{z+1}) \cdot v_{z+1},\cdots, \phi^{l}(p_z,c_n) \cdot v_n,  \nonumber\\
& \qquad \qquad \qquad  \phi^{s}(p_0,p_1),\cdots, \phi^{s}(p_0,p_z), v_{z+1}\cdot v_{z+1+m},...,v_{n-m}\cdot v_{n} ,\phi^r(C^r)).
\end{align}
\end{small}
\begin{figure}[!htb]
\centering
\epsfig{figure=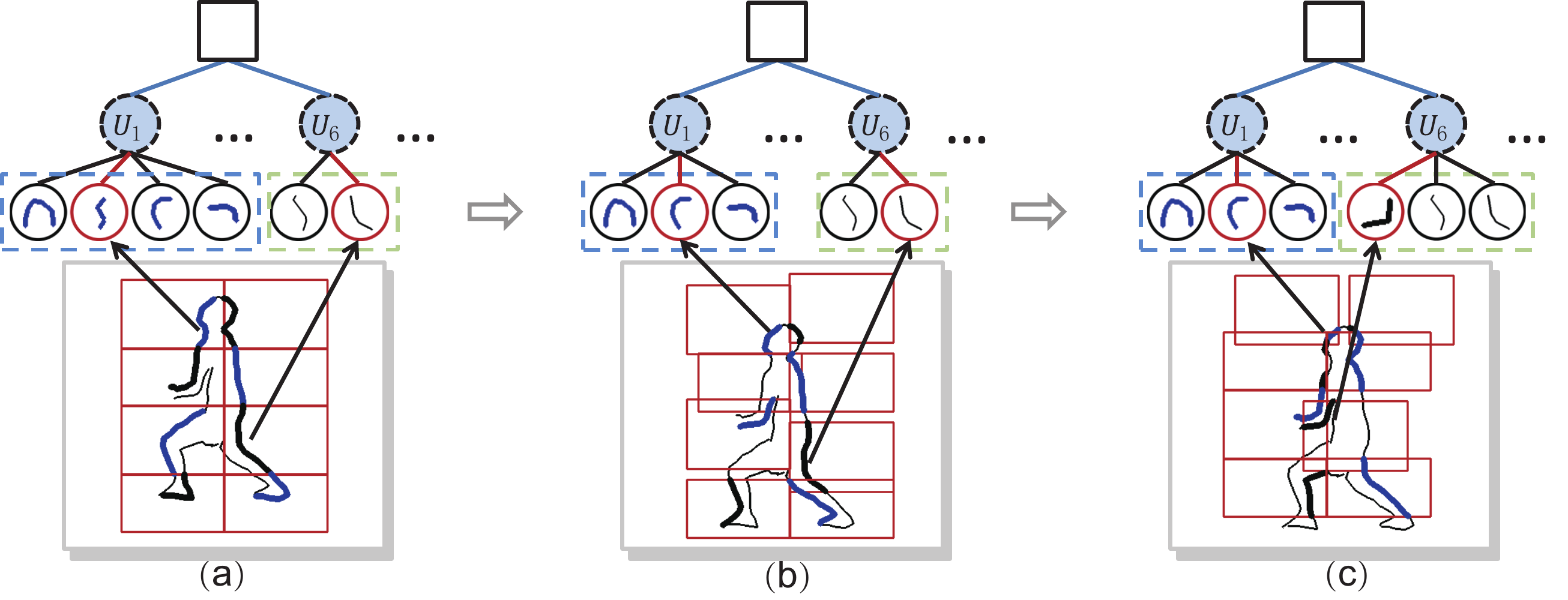,width=0.70\textwidth}
\vspace{-2ex}
\caption{Illustration of dynamical structure learning. Parts of the model, two or-nodes ($U_1,U_6$), are visualized in three intermediate steps. (a) The initial structure, i.e., the regular layout of an object. Two new structures are dynamically generated during iteration. (b) A leaf-node associated with $U_1$ is removed. (c) A new leaf-node is created and assigned to $U_6$.
 }\label{fig:structure}
\end{figure}

\vspace{-5ex}
\section{Inference}
\vspace{-2ex}
The inference task is to localize the optimal contour fragments within the detection window, which is slidden at all scales and positions of the edge map $X$. Assuming the root-node is located at $p_0$, the object shape is localized by maximizing $\mathcal{R}_{G}(X, H)$ defined in (\ref{eq:AOG_score}):
\begin{eqnarray}\label{eq:AOG score 2}
&& S(p_0, X) = \max_H \mathcal{R}_{G}(X, H).
\end{eqnarray}
\vspace{-4ex}

The inference procedure integrates the bottom-up testing and top-down verification:

\textbf{Bottom-up testing:} For each or-node $U_i$, its children leaf-nodes (i.e. the local classifiers) are utilized to detect contour fragments within the edge map $X$.  Assume that leaf-node $L_j, j\in ch(i)$ associated with $U_i$ is activated, $v_j = 1$, and the optimal contour fragment $c_j$ is localized by maximizing the response in Eq.(\ref{eq:or-node_score}), where the optimal location $p_{i,j}^{*}$ is also determined. Then we generate a set of candidates for each or-node, $\{c_j, p_{i,j}^{*}\}$, each of which is one detected contour fragments via the leaf-nodes. These sets of candidates will be passed to the top-down step where the leaf-node activation $\textbf{v}_{i}$ for $U_i$ can be further validated. We calculate the response for the bottom-up step, as,

\vspace{-4ex}
\begin{align}
\qquad \mathcal{R}_{bot}(V) = \sum_{i=1}^{z} \mathcal{R}_{U_i}(X,p_0,p_i^*,\textbf{v}_{i}),
\end{align}
\vspace{-3ex}

where $V = \{\textbf{v}_{i}\}$ denotes a hypothesis of leaf-node activation for all or-nodes. In practice, we can further prune the candidate contours by setting a threshold on $\mathcal{R}_{bot}(V)$. Thus, given the $V = \{\textbf{v}_{i}\}$, we can select an ensemble of contours $C^r=\{c_1,...,c_z\}$, each of which is detected by an activated leaf-node, $L_j, v_j = 1$.

\textbf{Top-down verification:} Given the ensemble of contours $C^r$, we then apply the global classifier at the root-node to verify $C^r$ by Eq. (\ref{eq:root_score}), as well as the accumulated pairwise potentials on the collaborative edges defined in Eq.(\ref{eq:CRF score}).

By incorporating the bottom-up and top-down steps, we obtain the response of  And-Or graph model by Eq.(\ref{eq:AOG_score}). The final detection is acquired by selecting the maximum score in Eq.(\ref{eq:AOG score 2}).

%\section{Learning and Inference}
\vspace{-1mm}
\section{Discriminative Learning for And-Or Graph}
\vspace{-2ex}

We formulate the learning of And-Or graph model as a joint optimization task for model structure and parameters, which can be solved by an iterative method extended from the CCCP framework~\cite{SVMICML2009}. This algorithm iterates to determine the And-Or graph structure in a dynamical manner: given the inferred latent variables $H = (P,V)$ in each step, the leaf-nodes can be automatically created or removed to generate a new structural configuration. To be specific, a new leaf-node is encouraged to be created as the local detector for contours that cannot be handled by the current model(Fig.~\ref{fig:structure}(c)); a leaf-node is encourage to be removed if it has similar discriminative ability as other ones(Fig.~\ref{fig:structure}(b)). We thus call this procedure dynamical CCCP (dCCCP).

%The learning method for our And-Or graph model is a iterative procedure integrating structure learning and parameter learning, which is an extension of the CCCP presented in ~\cite{SVMICML2009}. During the iterations, the leaf-nodes can be created or removed automatically to fit the inferred latent variables. To be specific, a new leaf-node is encouraged to be created as the local detector for contours that cannot be handled by the current model(Fig.~\ref{fig:structure}(c)); a leaf-node is encourage to be removed if it has similar discriminative ability as other ones(Fig.~\ref{fig:structure}(b)).

\vspace{-1mm}
\subsection{Optimization Formulation}
\vspace{-2ex}
Suppose a set of positive and negative training samples $(X_1,y_1)$,...,$(X_N,y_N)$ are given, where $X$ is the edge map, $y=\pm 1$ is the label to indicate positive and negative samples. We assume the samples indexed from $1$ to $K$ are the positive samples, and the feature vector for each sample $(X,y)$ as,

\vspace{-4ex}
\begin{equation}\label{eq:feature_phi}
 \qquad \phi(X,y,H) =
  \left\{
   \begin{array}{lr}
   \phi(X,H) & \mbox{if } y = +1 \\
   0 & \mbox{if } y = -1\\
   \end{array}
  \right.,
\end{equation}
\vspace{-3ex}

where $H$ is the latent variables. Thus, Eq.(\ref{eq:AOG score 2}) can be rewritten as a discriminative function,

\vspace{-4.5ex}
\begin{eqnarray} \label{eq:discriminative_fun}
&& S_{\omega}(X) = argmax_{y,H} (\omega \cdot \phi(X,y,H)).
\end{eqnarray}
\vspace{-4.5ex}

The optimization of this function can be solved by using structural SVM with latent variables,

\vspace{-4.5ex}
\begin{small}
\begin{align} \label{eq:learn_opt}
& \min_{\omega} \frac{1}{2} \| \omega \|^2  +  D\sum_{k=1}^N[\max_{y,H}(\omega \cdot \phi(X_k,y,H) + \mathcal{L}(y_k,y,H)) - \max_H (\omega \cdot \phi(X_k,y_k,H))],
\end{align}
\end{small}
\vspace{-3ex}

where $D$ is a penalty parameter(set as 0.005 empirically), and $\mathcal{L}(y_k,y,H)$ is the loss function. We define that $\mathcal{L}(y_k,y,H) = 0$ if $y_k = y$, ``1'' if $y_k \neq y$ in our method.

The optimization target in Equation(\ref{eq:learn_opt}) is non-convex. The CCCP framework~\cite{CCCP} was recently utilized in~\cite{SVMICML2009,LeoCCCP} to provide a local optimum solution by iteratively solving the latent variables $H$ and the model parameter $\omega$. However, the CCCP does not address the or-nodes in hierarchy, i.e., assuming the configuration of structure is fixed. In the following, we propose the dCCCP by embedding a structural reconfiguration step.

\vspace{-1mm}
\subsection{Optimization with dynamic CCCP}
\vspace{-2ex}

Following the original CCCP framework, we convert the function in Eq. (\ref{eq:learn_opt}) into a convex and concave form as,

\vspace{-4.5ex}
\begin{small}
\mathindent=0pt
\begin{align}\label{eq:cccp_f}
&\quad \min_{\omega}[ \frac{1}{2} \|\omega \|^2 + D \sum_{k=1}^N \max_{y,H}(\omega \cdot \phi(X_k,y,H) + \mathcal{L}(y_k,y,H))]  -  [D \sum_{k=1}^N \max_{H} (\omega \cdot \phi(X_k,y_k,H))] \\ \label{eq:opt_target}
& =  \min_{\omega} [f(\omega) - g(\omega)],
\end{align}
\mathindent=\leftmargini
\end{small}
\vspace{-3ex}

where $f(\omega)$ represents the first two terms, and $g(\omega)$ represents the last term in (\ref{eq:cccp_f}).

The original CCCP includes two iterative steps: (I) fixing the model parameters, estimate the latent variables $H^*$ for each positive samples; (II) compute the model parameters by the traditional structural SVM method. In our method, besides the inferred $H^*$, we need to further determine the graph configuration, i.e. the production of leaf-nodes associated with or-nodes, to obtain the complete structure. Thus, we insert one step between two original ones to perform the structure reconfiguration. The three iterative steps are presented as follows.

%During structure re-organization, we maintain the principle components of $\phi(X,y,H^*)$ for all samples obtained in step(i). According to ~\cite{DimRedPCA}, by applying PCA, we can represent a high-dimensional vector with a lower-dimensional one without significant loss of information. Thus the dominant patterns of $\phi(X,y,H^*)$ are well preserved after the adjustment. Since the hyperplane constructed in step(i) is in an addictive form of all $\phi(X,y,H^*)$, we can assume its consistency after structure re-organization, which ensure the convergence of our learning algorithm. The proposed dCCCP framework iterates with the following three steps.

{\bf (I)} For optimization, we first find a hyperplane $q_t$ to upper bound the concave part $-g(\omega)$ in Eq.(\ref{eq:opt_target}),

\vspace{-4ex}
\begin{equation}\label{eq:CCCP_cons}
 -g(\omega) \leq -g(\omega_t) + (\omega-\omega_t) \cdot q_t, \forall \omega.
\end{equation}
\vspace{-4ex}

where $\omega_t$ includes the model parameters obtained in the previous iteration. We construct $q_t$ by calculating the optimal latent variables $H_k^* = argmax_{H} (\omega_{t} \cdot \phi(X_k,y_k,H))$. Since $\phi(X_k,y_k,H) = 0$ when $ y_k = -1$, we only take the positive training samples into account during computation. Then the hyperplane is constructed as $q_t = - D\sum_{k=1}^N \phi(X_k,y_k,H_k^*)$.

{\bf (II)} In this step, we adjust the model structure by reconfiguring the leaf-nodes. In our model, each leaf-node is mapped to several feature dimensions of the vector $\phi(X,y,H^*)$. Thus, the process of reconfiguration is equivalent to reorganizing the feature vector $\phi(X,y,H^*)$. Accordingly, the hyperplane $q_t$ would change with $\phi(X,y,H^*)$, and would lead to non-convergence of learning. Therefore, we operate on $\phi(X,y,H^*)$ guided by the Principal Component Analysis(PCA). That is, we allow the adjustment only with the non-principal components (dimensions) of $\phi(X,y,H^*)$, in terms of preserving the significant information of $\phi(X,y,H^*)$~\cite{DimRedPCA}. As a result, $q_t$ is assumed to be unaltered. This step of model reconfiguration can be then divided into two sub-steps.
%We divide this step of structure learning into two sub-steps including: feature selection guided by PCA and structural clustering. Note that the reconfiguration is performed independently for each or-node. For convenience, we simplify the notation as: $\phi$ denotes $\phi(X_k,y_k,H_k^{*})$, $\phi^l$ denotes $\phi^l(p_i^k, c_i^k)$ and $\phi^{\prime}$ denotes $\phi^{\prime}(p_i^k, c_i^k)$ which will be defined below.
\begin{figure}[!htb]
\centering
\epsfig{figure=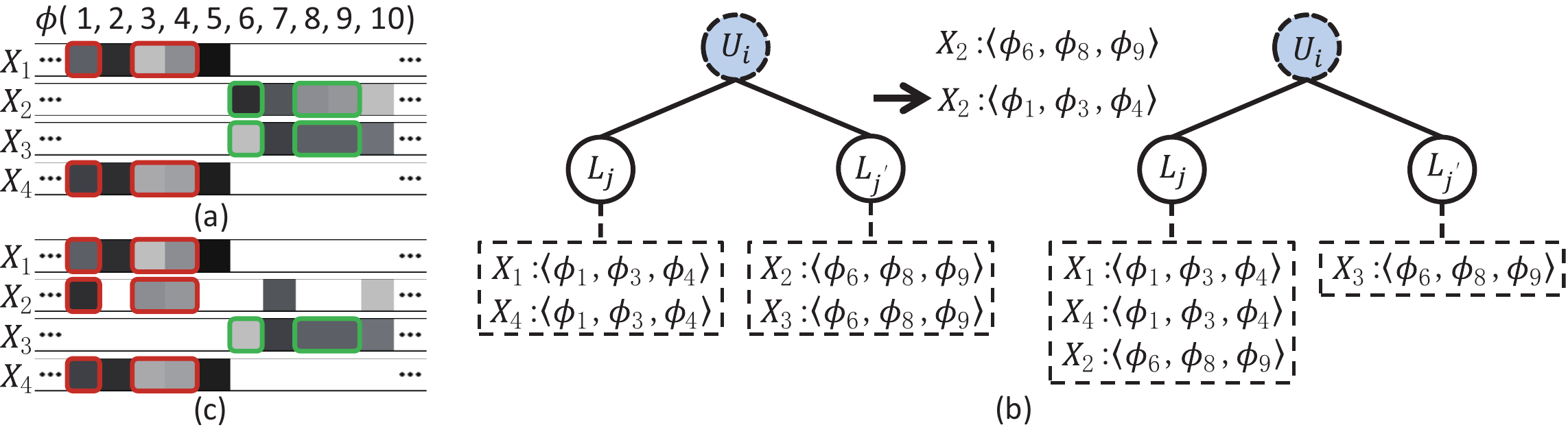,width=0.85\textwidth}
\vspace{-2ex}
\caption{A toy example for structural clustering. We consider $4$ samples, $X_1, \ldots, X_4$, for training the structure of $U_i$. (a) shows the feature vectors $\phi$ of the samples associated with $U_i$, and the intensity of the feature bin indicates the feature value. The red and green bounding boxes on the vectors indicate the non-principal features representing the detected contour fragments via two different leaf-nodes. (b) illustrates the clustering performed with $\phi^{\prime}$. The vector $\langle \phi_6,\phi_8,\phi_9 \rangle$ of $X_2$ is grouped from the right cluster to the left one. (c) shows the adjusted feature vectors according to the clustering. Note that clustering would result in structural reconfiguration, as we discuss in the text. This figure is encouraged to be view in electronic version. }\label{fig:step2}
\end{figure}

{\bf (i) Feature refactoring guided by PCA.} Given $\phi(X_k,y_k,H_k^*)$ of all positive samples, we apply PCA on them,

\vspace{-5ex}
\begin{align} \label{eq:pca_phi}
\phi(X_k,y_k,H_k^{*}) \approx u + \sum_{i=1}^{\mathcal{K}} \beta_{k,i} e_i,
\end{align}
\vspace{-3.5ex}

where $\mathcal{K}$ is the number of the eigenvectors, $e_i$ the eigenvector with its parameter $\beta_{k,i}$. We set $\mathcal{K}$ a large number so that $||\phi(X_k,y_k,H_k^{*}) - (u + \sum_{i=1}^{\mathcal{K}} \beta_{k,i} e_i)||_2 < \sigma$, $\forall k$. For the $jth$ bin of the feature vector, we consider it non-principal only if $e_{i,j} < \delta$ and $u_{j} < \delta$ for all $e_i$ and $u$, ($\sigma=2.0$, $\delta=0.001$ in experiments).

%irrelevant to the principle components if for all $e_i$ and $u$, the $jth$ bins of them have $e_{i,j} < \delta$ and $u_{j} < \delta$.($\sigma=2.0$, $\delta=0.001$ in the experiments)

For each or-node $U_i$, a set of detected contour fragments, $\{c_i^{1},c_i^{2},...,c_i^{K}\}$, are obtained with the given $H_k^*$ of all positive samples. The feature vectors for these contours that are generated by the leaf-nodes, $\{\phi^l(p_i^1, c_i^1),...,\phi^l(p_i^K, c_i^K)\}$, are mapped to different parts of the complete feature vector, $\{\phi(X_1,y_1,H_1^{*}),...,\phi(X_K,y_K,H_K^{*})\}$. More specifically, once we select the $jth$ bin for the all feature vectors $\phi^l$, it can be either principal or not in different vectors $\phi$. For all feature vector $\phi^l$, we select the non-principal bins to form a new vector. We thus refactor the feature vectors of these contours as $\{\phi^{\prime}(p_i^1, c_i^1),...,\phi^{\prime}(p_i^K, c_i^K)\}$.

%Since these contours are detected via different leaf-node detectors, their shape feature vectors $\{\phi^l(p_i^1, c_i^1),...,\phi^l(p_i^K, c_i^K)\}$ are associated to different parts of $\{\phi(X_1,y_1,H_1^{*}),...,\phi(X_K,y_K,H_K^{*})\}$. Thus for the $jth$ bins of all the $\phi^l$, some might be related to the principle components, the others are not. We select the bins that are irrelevant to the principle components in all $\phi^l$ to form a new set of feature vectors $\{\phi^{\prime}(p_i^1, c_i^1),...,\phi^{\prime}(p_i^K, c_i^K)\}$, representing the activated contours.

{\bf (ii) Structural reconfiguration by clustering.} To trigger the structural reconfiguration, for each or-node $U_i$, we perform the clustering for detected contour fragments represented by the newly formed feature vectors. We first group the contours detected by the same leaf-node into the same cluster as a temporary partition. Then the re-clustering is performed by applying the ISODATA algorithm and the Euclidean distance. And the close contours are grouped into the same cluster. According to the new partition, we can re-organize the feature vectors, i.e. represent the similar contour with the same bins in the complete feature vector $\phi$. Please recall that the vector of one contour is part of $\phi$. We present a toy example for illustration in Fig.~\ref{fig:step2}. The selected feature vector (non-principal) $\phi^{\prime}(p_i^2, c_i^2)=\langle \phi_6,\phi_8,\phi_9 \rangle$ of $X_2$ is grouped from one cluster to another; by comparing (a) with (c) we can observe that $\langle \phi_6,\phi_8,\phi_9 \rangle$ is moved to $\langle \phi_1,\phi_3,\phi_4 \rangle$.

With the re-organization of feature vectors, we can accordingly reconfigure the leaf-nodes corresponding to the clusters of contours. There are two typical states.

\vspace{-1ex}
\begin{itemize}
    \item New leaf-nodes are created once more clusters are generated than previous. Their parameters can be learned based on the feature vectors of contours within the clusters.
    \item One leaf-node is removed when the feature bins related to it are zero, which implies the contours detected by the leaf-node are grouped to another cluster.
\end{itemize}
\vspace{-1ex}

In practice, we constrain the extent of structural reconfiguration, i.e., only  few leaf-nodes can be created or removed for each or-node per iteration. After the structural reconfiguration, we denote all the feature vectors $\phi(X_k,y_k,H_k^{*})$ are adjusted to $\phi^{d}(X_k,y_k,H_k^{*})$. Then the new hyperplane is generated as $q_t^{d} = - D\sum_{k=1}^N \phi^{d}(X_k,y_k,H_k^{*})$.

%(1) The new leaf-nodes are created when the number of the clusters after re-clustering is larger than the number of the original leaf-nodes. Each cluster of contours is a ``potential'' leaf-node. The parameters of the new leaf-nodes will be learned based on $\phi^{\prime}$ of the contours in the corresponding clusters. Depending on the needs for handling the intra-class variance, the new leaf-nodes will be remained or removed in the next iterations.

%If the new leaf-nodes are really in need for handling the intra-class variance, they will be remained in the next iterations. Otherwise, they will be removed, described as follow.

%(2) The leaf-nodes are removed when the feature bins related to them in all $\phi$ are zero. It often happens when these leaf-nodes function little during detection in step(i), i.e., most of the feature bins related to them are already zero before clustering. Thus, these leaf-nodes are redundant and should be removed.

% constrain that there is at most one leaf-node created or removed for each $U_i$ per iteration. Although we use feature $\phi^{\prime}$ to represent the activated contours during clustering, the dominant parts of them are well preserved by fixing the principle components. Thus, we do not need to rectify the feature vector corresponding to the collaborative edges $(v_{z+1}\cdot v_{z+1+m},...,v_{n-m}\cdot v_{n})$.

\begin{figure}[!htb]
\centering
\epsfig{figure=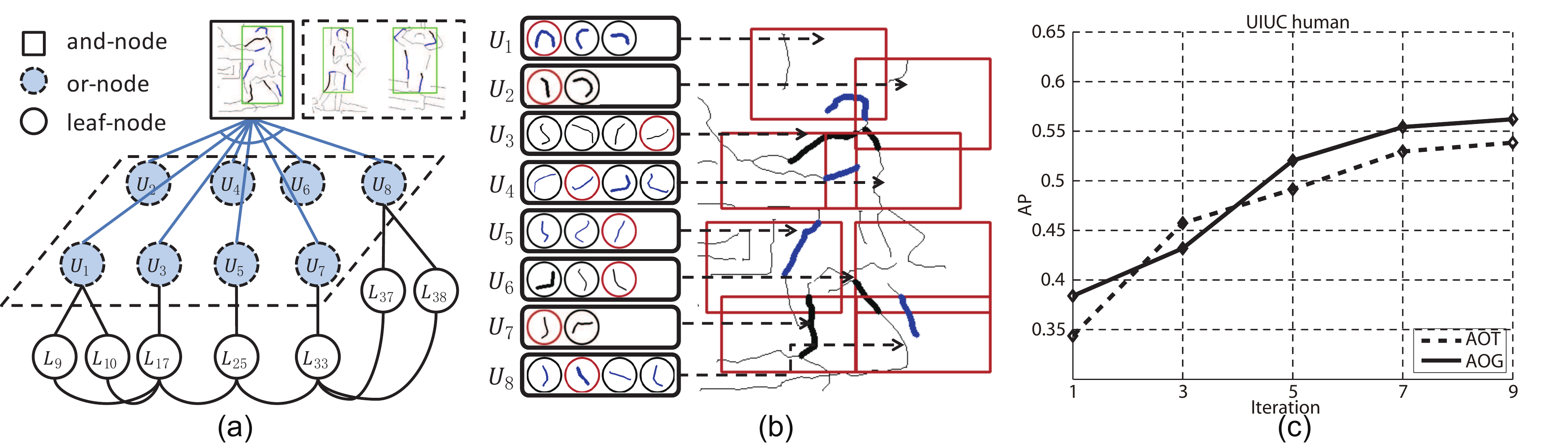,width=0.85\textwidth}
\vspace{-2ex}
\caption{ The trained And-Or graph model with the UIUC-People dataset. (a) visualizes the three layer model, where the images on the top imply the verification via the root-node. (b) exhibits the leaf-nodes associated with the or-nodes, $U_1, \ldots, U_8$; a practical detection with the activated leaf-nodes are highlighted by red. (c) shows the average precisions (AP) results generated by the And-Or tree (AOT) model and the And-Or graph (AOG) model. }
\label{fig:exp1}
\end{figure}

{\bf (III)} Given the newly generated model structures represented by the feature vectors $\phi^d(X_k,y_k,H_k^{*})$, we can learn the model parameters by solving $\omega_{t+1} = argmin_\omega[f(\omega)+\omega \cdot q_t^{d}]$. By substituting $-g(\omega)$ with the upper bound hyperplane $q_t^{d}$, the optimization task in Eq. (\ref{eq:cccp_f}) can be rewritten as,

\vspace{-3.5ex}
\begin{small}
\begin{align}
& \min_\omega \frac{1}{2} \|\omega\|^2 +  D\sum_{k=1}^N[\max_{y,H}(\omega \cdot \phi(X_k,y,H) + \mathcal{L}(y_k,y,H)) -  \omega \cdot \phi^d(X_k,y_k,H_k^{*})].
\end{align}
\end{small}
\vspace{-3ex}

This is a standard structural SVM problem, whose solution is presented as,

\vspace{-4.5ex}
\begin{eqnarray}
&& \omega^* = D \sum_{k,y,H} \alpha_{k,y,H}^* \Delta\phi(X_k,y,H),
\end{eqnarray}
\vspace{-3ex}

where $\Delta\phi(X_k,y,H) = \phi^d(X_k,y_k,H_k^{*}) - \phi(X_k,y,H)$. We calculate $\alpha^*$ by maximizing the dual function:

\vspace{-3.5ex}
\begin{small}
\mathindent=0pt
\begin{align}
& \quad \max_{\alpha} \sum_{k,y,H} \alpha_{k,y,H} \mathcal{L}(y_k,y,H) - \frac{ D }{2}\sum_{k,k^{\prime}}\ \ \sum_{y,H,y^{\prime},H^{\prime}} \alpha_{k,y,H} \alpha_{k^{\prime},y^{\prime},H^{\prime}} \Delta\phi(X_k,y,H)
\Delta\phi(X_{k^{\prime}},y^{\prime},H^{\prime}).
\end{align}
\mathindent=\leftmargini
\end{small}
\vspace{-3ex}

It is a dual problem in standard SVM, which can be solved by applying the cutting plane method~\cite{CuttingPlane} and Sequential Minimal Optimization~\cite{SMO}. Thus, we obtain the updated parameters $\omega_{t+1}$, and continue the 3-step iteration until the function in Eq.(\ref{eq:opt_target}) converges.

\vspace{-1mm}
\subsection{Initialization}
\vspace{-2ex}
At the beginning of learning, the And-Or graph model can be initialized as follows. For each training sample (whose contours have been extracted), we partition it into a regular layout of several blocks, each of which corresponds to one or-node. The contours fallen into the block are treated as the input for learning. Once there are more than two contours in one block, we select the one with largest length. Then the leaf-nodes are generated by clustering the selected contours without any constraints, and we can thus obtain the initial feature vector $\phi^d$ for each sample.

\vspace{-1mm}
\section{Experiments}
\vspace{-2ex}
We evaluate our method for object shape detection, using three benchmark datasets: the UIUC-People~\cite{UIUCHuman}, the ETHZ-Shape~\cite{PAS} and the INRIA-Horse~\cite{PAS}.

{\em Implementation setting.} We fix the number of or-nodes in the And-Or model as $8$ for the UIUC-People dataset, and $6$ in other experiments. The initial layout is a regular partition (e.g. $4 \times 2$ blocks for the UIUC-People dataset and $2 \times 3$ for others). There are at most $m=4$ leaf-nodes for each or-node. For positive samples, we extract their clutter-free object contours; for negative samples, we compute their edge maps by using the Pb edge detector~\cite{PbDetector} with an edge link method. The convergence of our learning algorithm take $6 \sim 9$ iterations. During detection, the edge maps of test images are extracted as for negative training samples, within which the object is searched at 6 different scales, 2 per octave. For each contour as the input to the leaf-node, we sample $20$ points and compute the Shape Context descriptor for each point; the descriptor is quantized with $6$ polar angles and $2$ radial bins. We adopt the testing criterion defined in the PASCAL VOC challenge: a detection is counted as correct if the intersection over union with the groundtruth is at least $50\%$.

{\bf Experiment I.} The UIUC-People dataset contains 593 images (346 for training, 247 for testing). Most of the images contain one person playing badminton. Fig.~\ref{fig:exp1}(b) shows the trained And-Or model(AOG) in that each of the $8$ or-nodes associates with $2 \sim 4$ leaf-nodes. To evaluate the benefit from the collaborative edges, we degenerate our model to the And-Or Tree (AOT) by removing the collaborative edges. As Fig.~\ref{fig:exp1}(c) illustrates, the average precisions (AP) of detection by applying AOG and AOT are $56.20\%$and $53.84\%$ respectively. Then we compare our model with the state-of-the-art detectors in ~\cite{HierarchicalPoslets,HumanExp1,HumanExp2,LatentSVM}, some of which used manually labeled models. Following the metric mentioned in ~\cite{HierarchicalPoslets}, to calculate the detection accuracy, we only consider the detection with the highest score on an image for all the methods. As Table.~\ref{table:1(a)} reports, our methods outperforms other approaches.
\begin{figure}[!htb]
\centering
\epsfig{figure=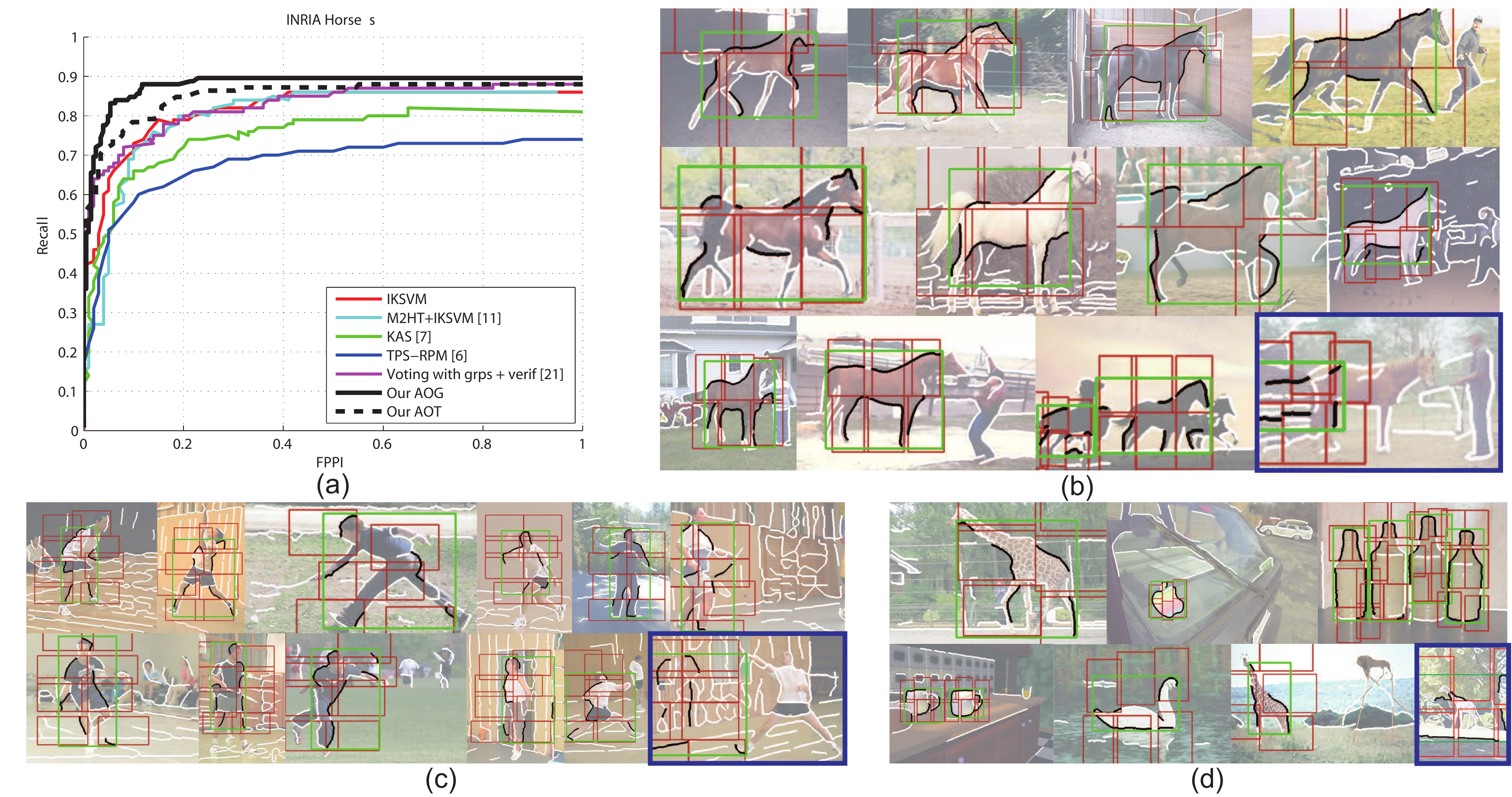,width=0.9\textwidth}
\vspace{-2ex}
\caption{(a)Experimental results with the recall-FPPI measurement on the INRIA-Horse database. (b),(c) and (d) shows a few object shape detections by applying our method on the three datasets, and the false positives are annotated by blue frames. }\label{fig:Horse_fppi}
\end{figure}

{\bf Experiment II.} The INRIA-Horse dataset consists of $170$ horse images and $170$ images without horses. Among them, $50$ positive examples and $80$ negative examples are used for training and remaining $210$ images for testing. Fig.~\ref{fig:Horse_fppi} reports the plots of false positives per image (FPPI) vs. recall. It is shown that our system substantially outperforms the recent methods: the AOG and AOT models achieve detection rates of $89.6\%$ and $88.0\%$ at $1.0$ FPPI, respectively; in contrast, the results of competing methods are: $87.3\%$ in ~\cite{VotingECCV2010}, $85.27\%$ in~\cite{MalikCVPR2009}, $80.77\%$ in~\cite{PAS}, and $73.75\%$ in~\cite{FerrariIJCV09}.

\begin{table}	 \small \addtolength{\tabcolsep}{-6pt}
	\begin{subtable}[t]{0.29\textwidth}
		\begin{tabular*}{\textwidth}{  p{2.8cm}p{1.1cm} l c } \toprule
                 & Accuracy \\
        \hline
        Our AOG                                     & \textbf{0.680} \\
        Our AOT                                     & 0.660 \\
        Wang et al.~\cite{HierarchicalPoslets}     & 0.668 \\
        Andriluka et al.~\cite{HumanExp1}                      & 0.506  \\
        Felz et al.~\cite{LatentSVM}                      & 0.486  \\
        Bourdev et al.~\cite{HumanExp2}                      & 0.458  \\ \bottomrule
        \end{tabular*}
		\caption{}\label{table:1(a)}
	\end{subtable}
\quad
	\begin{subtable}[t]{0.68\textwidth}
		\begin{tabular*}{\textwidth}{ p{2.8cm}p{1.52cm}p{1.05cm}p{1.1cm}p{1.01cm}p{1.01cm}p{1.01cm} l c c c c c c}
        \toprule
                                                    & Applelogos & Bottles & Giraffes & Mugs & Swans & Average\\
        \hline
        Our method                                  & \textbf{0.910} & 0.926 & \textbf{0.803} & 0.885 & \textbf{0.968} & \textbf{0.898} \\
        Ma et al.~\cite{LateckiCVPR2011}            & 0.881 & 0.920 & 0.756 & 0.868 & 0.959 & 0.877 \\
        Srinivasan et al.~\cite{ShiShapeCVPR2010}   & 0.845 & 0.916 & 0.787 & \textbf{0.888} & 0.922 & 0.872 \\
        Maji et al.~\cite{MalikCVPR2009}            & 0.869 & 0.724 & 0.742 & 0.806 & 0.716 & 0.771 \\
        Felz et al.~\cite{LatentSVM}                & 0.891 & \textbf{0.950} & 0.608 & 0.721 & 0.391 & 0.712 \\
        Lu et al.~\cite{ShapeGroup}                 & 0.844 & 0.641 & 0.617 & 0.643 & 0.798 & 0.709 \\
        \bottomrule
        \end{tabular*}
		\caption{}\label{table:1(b)}
	\end{subtable}
    \vspace{-2ex}
	\caption{(a) 
Comparisons of detection accuracies on the UIUC-People dataset.
            (b) Comparisons of average precision (AP) on the ETHZ-Shape dataset.}\label{table:1}
\end{table}

%From the results of shape detection, some of them are exhibited in Fig.~\ref{fig:detect_results} (b), the improvements are basically made by the accurate location in the context of (i) inconsistent shape contours (caused by pose variants or occlusions) and (ii) noisy edge maps.

{\bf Experiment III.} We test our method with more object categories on the ETHZ-Shape dataset: Applelogos, Bottles, Giraffes, Mugs and Swans. For each category (including $32 \sim 87$ images), half of the images are randomly selected as positive examples, and $70 \sim 90$ negative examples are obtained from the other categories as well as backgrounds. The trained model for each category is tested on the remaining images. Table ~\ref{table:1(b)} reports the results evaluated by the mean average precision. Compared with the current methods~\cite{MalikCVPR2009,ShiShapeCVPR2010,LatentSVM,ShapeGroup,LateckiCVPR2011}, our model achieves very competitive results.

A few results are visualized in Fig.\ref{fig:Horse_fppi}(b),(c) and (d) for experiment I, II, and III respectively.

\vspace{-2mm}
\section{Conclusion}
\vspace{-3ex}

This paper proposes a discriminative contour-based object model with the And-Or graph representation. This model can be trained in a dynamical manner that the model structure is automatically determined during iterations as well as the parameters. Our method achieves the state-of-art of object shape detection on challenging datasets.

%Moreover,
%
%
%In this paper, we propose a novel And-Or graph model, which can  capture  the large intra-class variance and the interplay between local parts, for object shape detection.
%We also present a new learning method to train the model parameters and structure simultaneously in a semi-supervised way.

%And it is worth to mention that our method is very general and can be applied to other tasks in computer vision.

%-------------------------------------------------------------------------

%{\small
%\bibliographystyle{ieee}
%
%\begin{thebibliography}
%
%
%
%\bibitem{Todorovic_ECCV2010}
%N. Payet and S. Todorovic, From a Set of Shape to Object Discoverty, In {\em ECCV}, 2010.
%
%
%
%
%\end{thebibliography}
%}

\bibliographystyle{ieee}

\end{document}